\documentclass[conference]{IEEEtran}
\IEEEoverridecommandlockouts
\usepackage{cite}
\usepackage{amsmath,amssymb,amsfonts}
\usepackage{algorithmic}
\usepackage{graphicx}
\usepackage{textcomp}
\usepackage[dvipsnames]{xcolor}

\usepackage{booktabs}  
\usepackage{graphicx} 
\usepackage{array}
\usepackage{cite}
\usepackage{multi row}
\usepackage{subcaption}
\usepackage{url}
\usepackage{color}
\usepackage{amsfonts}
\usepackage{amsmath}
\usepackage[ruled,vlined]{algorithm2e}
\usepackage[justification=centering, font=footnotesize]{caption}
\usepackage{verbatim}
\usepackage{tikz}
\DeclareMathOperator*{\argmax}{argmax}
\DeclareMathOperator*{\argmin}{argmin}
\usetikzlibrary{arrows.meta,positioning , fit, decorations.pathreplacing, shapes.geometric,backgrounds, hobby}
\usepackage[pscoord]{eso-pic}
\def\BibTeX{{\rm B\kern-.05em{\sc i\kern-.025em b}\kern-.08em
    T\kern-.1667em\lower.7ex\hbox{E}\kern-.125emX}}
    
\makeatletter
\newcommand{\newlineauthors}{%
  \end{@IEEEauthorhalign}\hfill\mbox{}\par
  \mbox{}\hfill\begin{@IEEEauthorhalign}
}
\makeatother
    
\begin{document}

\title{On the Robustness of Cooperative Multi-Agent Reinforcement Learning
}

\author{\IEEEauthorblockN{\fontdimen2\font=0.35ex{Jieyu Lin\IEEEauthorrefmark{1},
Kristina Dzeparoska\IEEEauthorrefmark{1}, 
Sai Qian Zhang\IEEEauthorrefmark{2},
Alberto Leon-Garcia\IEEEauthorrefmark{1},
Nicolas Papernot\IEEEauthorrefmark{1}\IEEEauthorrefmark{3}}} 
\IEEEauthorblockA{
\IEEEauthorrefmark{1}University of Toronto\\
\IEEEauthorrefmark{2}Harvard University\\
\IEEEauthorrefmark{3}Vector Institute\\
{\{jieyu.lin, kristina.dzeparoska\}@mail.utoronto.ca}
  }
}

\maketitle

\begin{abstract}
In cooperative multi-agent reinforcement learning (c-MARL), agents learn to cooperatively take actions as a team to maximize a total team reward. We analyze the robustness of c-MARL to adversaries capable of attacking one of the agents on a team. Through the ability to manipulate this agent's observations, the adversary seeks to decrease the total team reward. 

Attacking c-MARL is challenging for three reasons: first, it is difficult to estimate team rewards or how they are impacted by an agent mispredicting; second, models are non-differentiable; and third, the feature space is low-dimensional. Thus, we introduce a novel attack. The attacker first trains a policy network with reinforcement learning to find a wrong action it should encourage the victim agent to take. Then, the adversary uses targeted adversarial examples to force the victim to take this action. 

Our results on the StartCraft II multi-agent benchmark demonstrate that c-MARL teams are highly vulnerable to perturbations applied to one of their agent's observations. By attacking a single agent, our attack method has highly negative impact on the overall team reward, reducing it from 20 to 9.4. This results in the team's winning rate to go down from 98.9\% to 0\%. 

\end{abstract}


\section{Introduction}\label{sec:intro}

Algorithms for cooperative Multi-Agent Reinforcement Learning (c-MARL) are driving progress in a wide range of applications such as traffic light control\cite{wiering2000multi}, autonomous driving \cite{shalev2016safe}, and cellular base station control \cite{de2018cooperative}. Because they involve critical infrastructure, understanding the robustness of c-MARL agents to adversarial manipulations of their environment is a pre-requisite to their deployment in production. 

The goal of Reinforcement Learning (RL) is to optimize the actions taken by agents in complex environments, by learning actions that effectively maximize a long-term reward. 
Techniques for c-MARL were introduced to adapt RL algorithms to environments in which a team of multiple RL agents interact. Agents that form a team collaborate towards a common goal. 
c-MARL presents a number of key challenges including non-stationary environment, and limited bandwidth for information sharing internal to a team once training is completed and agents are deployed~\cite{zhang2019efficient}.
Combined with partial observability (any given agent may not always have access to all observations made by other agents in the team), it is more difficult for an agent in c-MARL to select the best action and maximize the total team reward.

RL agents are known to be vulnerable to adversaries perturbing their observations with adversarial examples~\cite{huang2017adversarial}, as well as adversaries directly controlling the actions of one of the victim's opponents~\cite{gleave2019adversarial}. 
When exploited, this vulnerability results in agents taking unintended actions, often with adverse consequences. 
We hypothesize that cooperative aspects of c-MARL agents, which enable them to outperform classic RL agents when operating as a team, also increase the vulnerability of a team to failures of one of its constituent agents. 

To the best of our knowledge, this work is the first to study c-MARL in adversarial settings and characterize the attack surface exposed by enabling agents to cooperate more effectively. Our goal is to understand the impact an adversary can have on the long-term reward of a team of agents. We consider an adversary who is capable of controlling the observations of a single \textit{victim agent}. The adversary mounts their attack by presenting the victim agent with adversarial examples. As c-MARL systems are increasingly deployed in critical infrastructure, there is a pressing need to assess their robustness in order to develop tailored defense mechanisms that promote security and safety in c-MARL. 

Several aspects of c-MARL preclude us from directly applying known attack algorithms for RL agents:

\begin{enumerate}
    
    \item \textbf{Team Reward Estimation \& Non-Differentiability}: Because joint value estimation networks used to estimate a team's total reward are trained with the assumption that agents will always choose the optimal action, they do not accurately predict the team reward when some of the agents misbehave (i.e., select non-optimal actions). Furthermore, estimating the team reward involves non-differentiable operations such as computing a maximum. This makes it difficult for an attacker to estimate (and thus decrease) the total team reward.

    \item \textbf{Misclassification $\boldsymbol{\neq}$ Reward}: High misclassification rates are not sufficient to lead to failure of RL. Wrong actions taken by agents at any point in time need to impact long-term rewards for the adversary to succeed.

    \item \textbf{Low-dimensional Input Feature Space}: In many c-MARL settings, agents are processing feature vectors with lower dimensions than images; hence, this limits the ways an adversary can perturb these observations. 

    \end{enumerate}

To address these challenges, we propose a novel two-step attack method (see Figure \ref{fig:main_diagram}). The first step uses RL to train an adversarial policy. This policy helps the adversary select actions which, if taken by an agent, would minimize the total team reward. This method only requires black-box access to the agent. The second step perturbs the agent's observation with an adversarial example so as to lure it to perform this adversarial action. To accommodate for the specificities of c-MARL, we refine existing approaches for finding targeted adversarial examples. In particular, we address the low input dimensionality and lack of smoothness of the optimization objective by adaptively setting hyperparameters of the attack. Finally, we introduce a regularization penalty to the training objective of our adversarial policy (the first step of the attack) to ensure that this first step produces a target adversarial action that is more easily achievable by the second step of the attack. This ensures that adversarial examples found can consistently decrease the reward and team win rate simultaneously. 

In our evaluation, we show that by perturbing the observation of a single agent, we can jeopardize the team almost as effectively as an adversary directly controlling this agent’s actions. This significantly reduces the total team reward of the c-MARL system. Our results on the StartCraft II multi-agent benchmark~\cite{samvelyan2019starcraft} show that an attacker able to perturb with an average of 8.3 L1 norm (out of 96 features ranging from $-1$ to 1) in the victim's input observation can successfully degrade the system’s team win rate from 98.9\% to 0\% (because its reward goes down from 20 to 9.4). 

Considering we were able to produce such results by attacking a single c-MARL agent, we emphasize in Section~\ref{sec:defenses} the necessity for suitable defenses, robust agents, and other protection measures against adversarial example attacks. Our contributions are listed as follows:
\begin{itemize}
    \item We propose a novel two-step attack for c-MARL system that is capable of reducing the total team reward by only perturbing the observation of a single agent. 
    \item We regularize the training of our adversarial policy to improve our two-step attack, demonstrating that reinforcement learning can be an effective adaptive strategy in the presence of difficult optimization landscapes when crafting adversarial examples (e.g., as is the case when defenses perform gradient masking~\cite{papernot2017practical,athalye2018obfuscated}).
    \item We extend existing adversarial example methods to create d-JSMA which is suitable for attacking RL model with low-dimensional feature space.
    \item We demonstrate the feasibility and effectiveness of our attack on c-MARL in the StarCraft II multi-agent benchmark. Our adversary brings down the team win rate from 98.9\% to 0\% by perturbing a single agent's observations.
\end{itemize}

\section{Problem Statement}\label{sec:back}

\subsection{Centralized Training Decentralized Execution and QMIX}

In c-MARL, due to the partial observability and limited communication among agents, learning decentralized policies is required to support decentralized execution. Current state-of-the-art MARL methods 
apply centralized training scheme, where agents can exchange their local information during training phase, and act independently during execution.

\begin{figure}[t]
  \centering
  \includegraphics[scale=0.55]{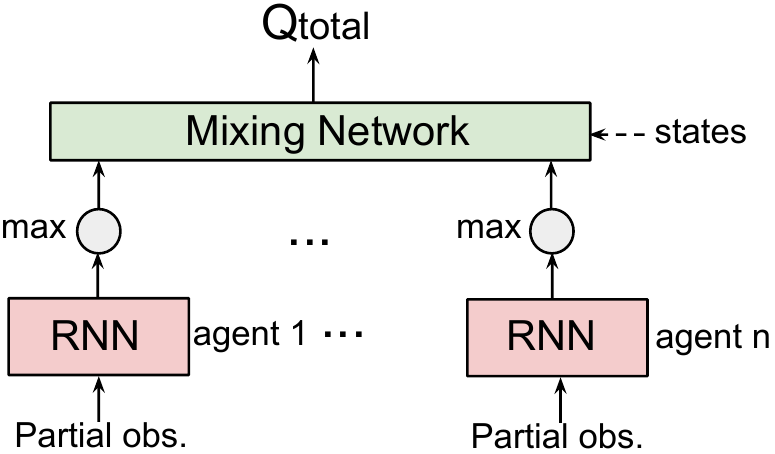}
  \caption{High-level architecture of the QMIX algorithm.}
  \label{fig:two}
\end{figure}

One approach for centralized training and decentralized execution is QMIX~\cite{rashid2018qmix} (Figure \ref{fig:two}), a multi-agent Deep Q-learning (DQN) algorithm, where each agent uses an RNN-based DQN network to estimate the action values based on their partial observation.
DQN has recently achieved success in a variety of RL tasks. It uses experience replay and a target network to improve stability and convergence of RL training. 
To maximize the total team reward, QMIX estimates the joint action values through a mixing network that takes in each agent's selected action $Q$-value (i.e., action with max $Q$-value) and the current state to estimate the total team reward, $Q_{total}$. With an accurate estimation of $Q_{total}$, we can fine-tune the individual agent's $Q$ network, so a greater total team reward can be achieved during execution. 

\subsection{Threat Model}\label{sec:threat_model}

We model the c-MARL game as a Dec-POMDP\cite{oliehoek2016concise} with $N$ agents cooperating to execute some tasks, with the goal of maximizing their total team reward. More specifically, at each time step $t$, each agent $i$ ($0 < i < N$) takes the (partial) observation $o_t^i$ as input, and performs an action $a_t^i$ with the highest action value. Let $\boldsymbol{a_t} = \{a^i_t\}$ be a vector of actions taken by all the agents at time $t$. The goal is to find the optimal $\boldsymbol{a_t}$ that maximizes the total team reward $R$.
 
Our threat model considers a single vulnerable agent $\hat{i}$ whose observation $o_t^{\hat{i}}$ can be modified by the adversary. We assume the adversary has already predetermined this agent. This is realistic, as in a real-world distributed system environment, an adversary can potentially gain access to a single node, but may not be able to affect all nodes. The adversary's goal is to minimize $R$ by making a minor modification to $o_t^{\hat{i}}$.

\subsection{Attack Overview}

\begin{figure}[t]
  \centering
  \includegraphics[width=\linewidth]{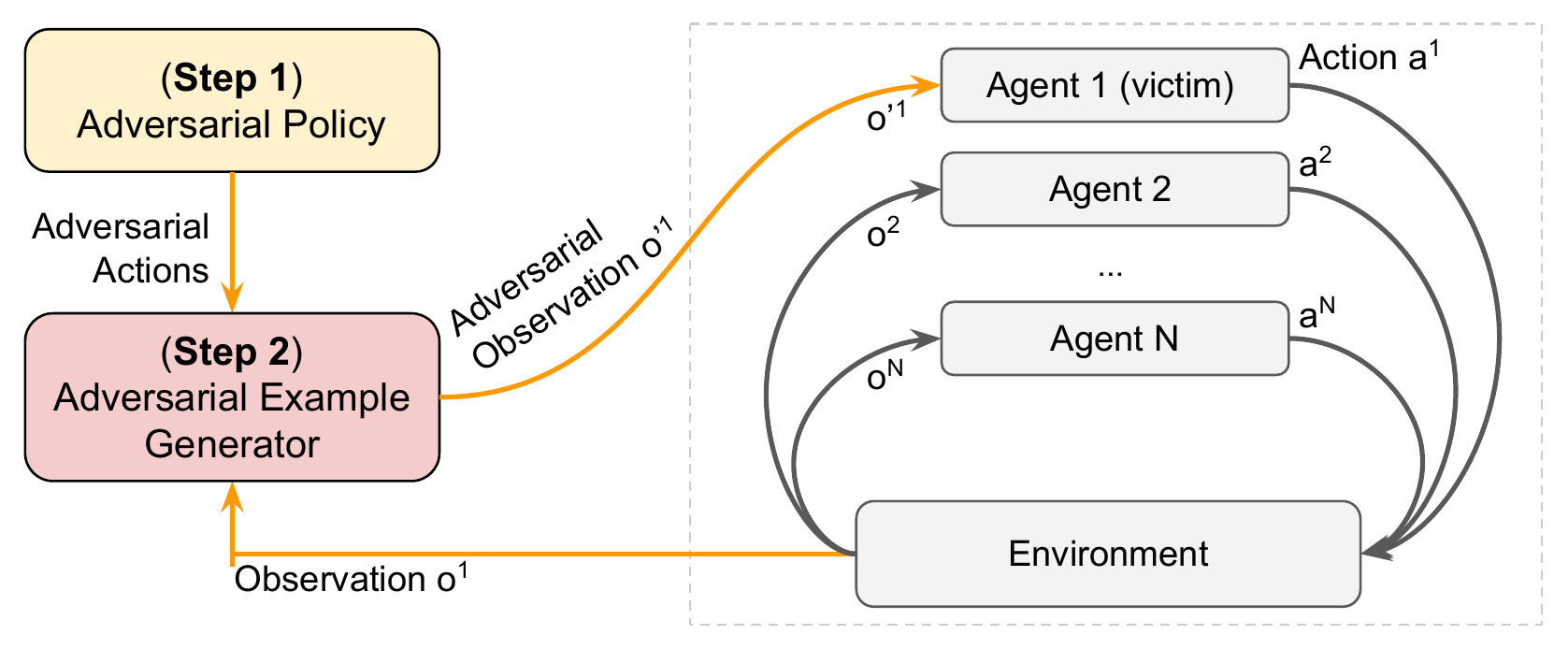}
  \caption{Two-step attack on a single agent in a c-MARL environment, as depicted by the orange lines and boxes.}
  \label{fig:main_diagram}
\end{figure}

Figure \ref{fig:main_diagram} shows a high-level diagram of our two-step attack. In the first step, we use RL to learn an adversarial policy that estimates which action, if taken by the victim, would result in the largest decrease in the total reward $R$. In the second step, we use gradient-based targeted adversarial example crafting to perturb the victim's observation such that it will take the action returned by the adversarial policy in the first step. Intuitively, using RL rather than a gradient-based strategy in the first step circumvents any non-differentiable operation found in how QMIX estimates the total reward $R$. Nevertheless, in the second step, the attacker can still apply gradient-based methods to craft adversarial examples---once it has identified the target action, these adversarial examples should conduce. 

Our two-step attack can be implemented in realistic threat models. The first step only assumes black-box access: adversaries query the black-box agent policy and its environment. The second step assumes the adversary has sufficient access to find adversarial examples. 
We instantiate a gradient-based strategy, which requires access to the victim's model parameters and observations. However, our attack could be extended to the black-box setting with finite-differences~\cite{chen2017zoo} or by creating a replica of the victim's model and leveraging the transferability property of adversarial examples. 

\section{Attack Methods}\label{sec:atatcks}

In this section, we discuss both steps of our attack in details.

\subsection{Learning adversarial policy} \label{sec:direct_control}
In this part of the attack, the goal is to optimize the behavior of the victim agent so its actions will result in a decreased total team reward $R$. We first explore na\"ive ways of selecting target actions for the victim agent:

\begin{itemize}
    \item \textbf{Random action}: Randomly selecting an available action.
    \item \textbf{Local worst action (LW)}: Selecting the action with the lowest Q-value based on the output of its Q network.
    \item \textbf{QMIX worst action}: Selecting the action leading to the min $Q_{total}$ value output by the mixing network. We omit the $max$ operation, and instead feed each available action's Q-value separately through the mixing network, and select the action leading to the worst $Q_{total}$.  
\end{itemize}

Despite the success of the above methods in reducing $R$, the victim agent's negative effect upon the team performance remains limited. To help improve the robustness of c-MARL, it is important to understand the worst-case behavior of a single agent. We propose RL-based optimization to more effectively select an action for the victim agent. We design two variants: 

\begin{itemize}
    \item \textbf{Optimized worst adversarial policy (OW)}: Using RL to learn a victim agent policy that minimizes $R$. 
    
    \item \textbf{OW with regularization (OWR)}: Adding regularization to OW, to be more efficiently combined with the adversarial example search performed in step two of the attack.

\end{itemize}

\textbf{OW}: To use RL to train an adversarial policy, we first turn to the formulation of c-MARL. Following the notations in Section~\ref{sec:threat_model}, the goal is to maximize the sum of discounted team rewards over time $t$:
 
\begin{equation}
    \argmax_{\theta^{1},\theta^{2},... \theta^{N}} \sum_{t=0}^{\infty}\gamma^{t}R_{t}(s_{t}, \boldsymbol{a_t}, s_{t+1}) \label{eqn:marl_r}
\end{equation}
where $\theta$ represents the parameters for each agent’s policy, $\gamma$ is the reward discount factor, $s_{t}$ is the state, and $s_{t+1}$ is determined by a transition function $T$ in the environment. The goal of centralized training is to maximize the total reward by tuning the parameters of all the agents' policy. 

In an adversarial environment, we can consider the policies of all the non-victim agents as fixed, i.e. deterministic, hence from the victim's perspective, they are considered as part of the environment. Thus, we can formulate our attack as a single agent RL problem, with the goal of minimizing $R$:
\begin{equation}
    \argmin_{\theta^{\prime v}} \sum_{t=0}^{\infty}\gamma^{t}R_{t}(s_{t}, a^{v}_{t}, s_{t+1}) \label{eqn:marl_adv}
\end{equation}
where $a_{t}^{v}$ is the victim agent's action at time step $t$, and $\theta^{\prime v}$ parameterizes the victim agent’s policy. With Equation~\ref{eqn:marl_adv}, we can train an adversarial policy that selects actions for the victim agent that will minimize $R$. In our implementation, we use DQN to train such an adversarial policy.

\textbf{OWR}: Despite OW's ability to learn an effective adversarial policy, this may not be sufficient for the adversary to succeed. Some of the target actions returned by the adversarial policy in this first step may be difficult to achieve given a limited perturbation budget when crafting an adversarial example for the victim agent in the second step. Thus, we propose OWR, which regularizes the objective optimized to train the adversarial policy. The new penalty reflects how difficult it would be to find an adversarial example for the predicted target.

Indeed, we observe that the adversarial example search sometimes fails to achieve a target action when the difference between the target action Q-value and the original action Q-value becomes too large. Intuitively, some actions are harder to achieve than others through adversarial examples because they are further away in the agent's decision space. To improve the attack success rate, we observe that the victim does not need to perform the most optimal action at each time step, instead it should focus on selecting the most optimal actions for the steps that have high negative impact on the team reward.

To incorporate this idea, we modify the DQN training used in OW to include a regularization term. The loss function of the attacker's Q network with regularization is defined as:
\begin{equation}
    L = \sum_{b=1}^{B}\sum_{t=1}^{T}[(y^b_{target} - Q_{target}(o^{v,b}_t, a_t^{v,b}))^2 + \lambda {d_{diff}}^2]
\end{equation}
where $y_{target}^b = r^b_t + \gamma max_{a_{t+1}}Q_{target}(o^{v,b}_{t+1}, a_{t+1}^{v,b})$, $r_t^b$ is the reward for time step $t$, $d_{diff}$ is the Q-value difference between the target action and the original best action, $\lambda$ is the weight of the $d_{diff}$ regularization, and $b$ represents the batch index.

\subsection{Adversarial example generation to lure victim agents}\label{sec:perturb}
\begin{figure}[t]
    \centering
    \begin{subfigure}{0.24\textwidth}
        \centering
        \includegraphics[width=\linewidth]{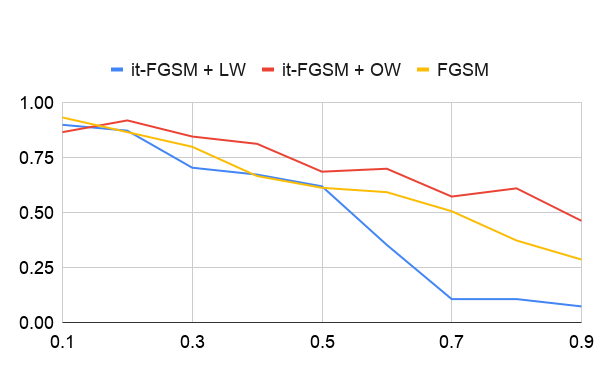}
        \caption{Team win rate}
        \label{subfig:fgsm_winrate}
    \end{subfigure}
    \begin{subfigure}{0.24\textwidth}
        \centering
        \includegraphics[width=\linewidth]{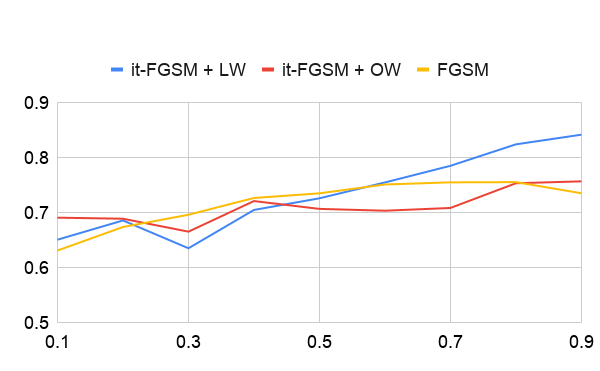}
        \caption{Misclassification Rate}
    \end{subfigure}
    \begin{subfigure}{0.24\textwidth}
        \centering
        \includegraphics[width=\linewidth]{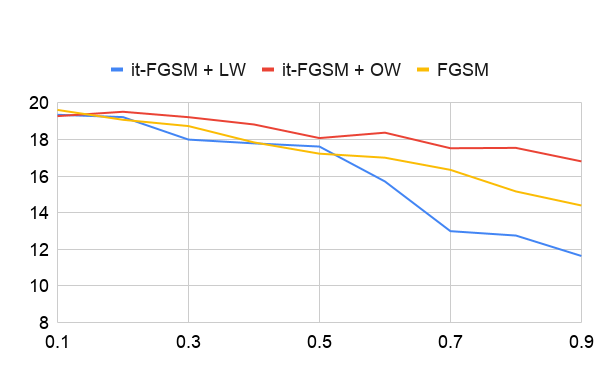}
        \caption{Total Reward\\~}
    \end{subfigure}
    \begin{subfigure}{0.22\textwidth}
        \centering
        \includegraphics[width=\linewidth]{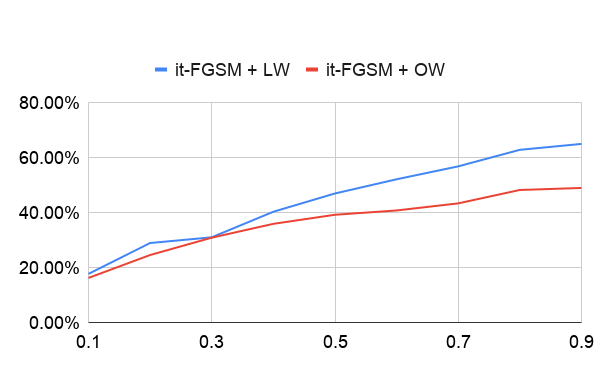}
        \caption{Target Action Success Rate}
    \end{subfigure}
    \caption{Performance of FGSM-based methods as a function of the perturbation budget $\epsilon$.}
    \label{fig:fgsm_eval}
\end{figure}

Next, the attack perturbs observations to have the victim select the action output by the adversarial policy from  Section~\ref{sec:direct_control}. One could hope to apply existing gradient-based adversarial example algorithms. However, the attack needs to be targeted. Indeed, we find in Figure~\ref{subfig:fgsm_winrate} that naively applying the FGSM~\cite{goodfellow2014explaining} results in total team rewards (28.7\%) that remain higher than what could be achieved by a random action baseline shown in Table~\ref{tab:1} (3.4\%). 
The attack also needs to retain sufficient flexibility to handle constraints placed on low-dimensional inputs observed from the agent's environment. 
This would require finding a differentiable loss function that hardcodes some of these constraints for the CW method~\cite{carlini2017towards} to be applicable. 

To more effectively craft adversarial examples for reducing $R$, we extended existing approaches to create two new methods: \textit{Iterative target-based FGSM method (it-FGSM)} and \textit{Dynamic budget JSMA attack (d-JSMA)}.

\textbf{it-FGSM}: We adapt the i-FGSM \cite{kurakin2016adversarial} algorithm to a targeted method by taking gradient on the target Q-value instead of the loss, that is changing $\nabla_X L(X)$ to $\nabla_X Q_{target}(X)$.

Figure \ref{fig:fgsm_eval} shows the performance of it-FGSM compared to FGSM. it-FGSM methods are more effective in reducing the total team reward and team win rate. it-FGSM + LW achieves the best performance by reducing the team reward and team winning rate to 11.63 and 7.33\%, respectively. However, we observe the performance of it-FGSM + OW is sub-optimal due to the lower target action success rate. Therefore, we conclude that the FGSM-based method is still not sufficiently effective for our problem setting. 

\begin{figure}[t]
  \centering
  \includegraphics[scale=0.45]{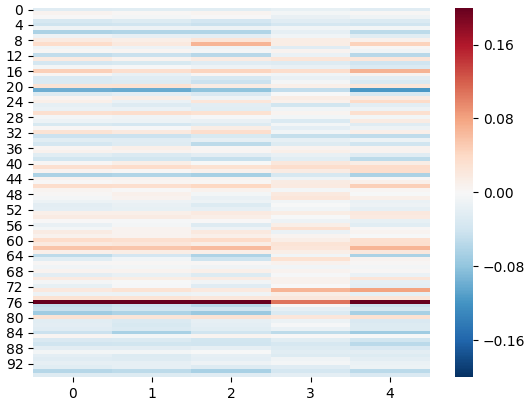}  
  \caption{Heat map of the gradient of action Q-values with respect to the input features. The x-axis denotes the available actions $a_i$, in this case $0 \leq i \leq 4$; y-axis denotes the observation features, $o_i$, $0 \leq i \leq 96$}
  \label{fig:heat}
\end{figure}

\textbf{d-JSMA}: JSMA~\cite{papernot2016limitations} is by default a targeted method, so it is suitable for our setting. It constructs a saliency map and perturbs the features with the highest saliency score (i.e. features that have large positive gradients for the target class and large negative sum of gradients for the non-target class). In our attempt to apply JSMA to our problem, we encountered two challenges, hence proposed d-JSMA to overcome them: 

1. \textbf{Two features bidirectional perturbation}: In our setting, when perturbing only a single feature, it is difficult to find features that have a positive gradient to our target action, but negative sum for the rest.  
Figure~\ref{fig:heat} shows an example gradient of action Q-values, where the gradients of different actions with respect to a single feature are often similar. Hence, perturbing a feature that increases the target action's Q-value inadvertently causes an increase to the rest of the actions (i.e. all Q-values change in the same direction). This is understandable in an RL setting, as changing a certain feature (e.g. victim's health has the highest gradient value in Figure~\ref{fig:heat}) may impact the final team reward regardless of the action. 

We address this issue by perturbing two features at the same time, by modifying Algorithm 3 proposed in~\cite{papernot2016limitations}. Formally, in d-JSMA, the attacker wants to perturb two input features simultaneously to increase the output Q-value of the target action $Q_{t}$ and decrease the output Q-values of the non-target actions, until the $\argmax(Q)$ selects the target action. We can accomplish this by perturbing the input features using the following saliency map  $\boldsymbol{S}(\boldsymbol{X}, t)$ and perturbation direction $d_{perturbation}(\boldsymbol{X}, t)$
\begin{equation}
\label{eqn:saliency}
\begin{split}
& \boldsymbol{S}(\boldsymbol{X}, t)[i, j] = \\ 
& \begin{cases}
    0 \text{ if } (\frac{\partial\boldsymbol{Q}_t(\boldsymbol{X})}{\partial\boldsymbol{X_i}} + \frac{\partial\boldsymbol{Q}_t(\boldsymbol{X})}{\partial\boldsymbol{X_j}})
    [\sum\limits_{k\neq t}(\frac{\partial\boldsymbol{Q}_k(\boldsymbol{X})}{\partial\boldsymbol{X_i}} + \frac{\partial\boldsymbol{Q}_k(\boldsymbol{X})}{\partial\boldsymbol{X_j}})] > 0\\ 
    -(\frac{\partial\boldsymbol{Q}_t(\boldsymbol{X})}{\partial\boldsymbol{X_i}} + \frac{\partial\boldsymbol{Q}_t(\boldsymbol{X})}{\partial\boldsymbol{X_j}})
    [\sum\limits_{k\neq t}(\frac{\partial\boldsymbol{Q}_k(\boldsymbol{X})}{\partial\boldsymbol{X_i}} + \frac{\partial\boldsymbol{Q}_k(\boldsymbol{X})}{\partial\boldsymbol{X_j}})] \text{ otherwise}
\end{cases}
\end{split}
\end{equation}

\begin{equation}
    d_{perturbation}(\boldsymbol{X}, t)[i, j] = Sign(\frac{\partial\boldsymbol{Q}_t(\boldsymbol{X})}{\partial\boldsymbol{X_i}} + \frac{\partial\boldsymbol{Q}_t(\boldsymbol{X})}{\partial\boldsymbol{X_j}})
\end{equation}

\noindent where $\boldsymbol{X}$ is the victim agent's observation, i, j are the indices of the input features. The condition in the first line of Equation~\ref{eqn:saliency} filters out feature pairs whose perturbation leads to Q-value changes in the same direction for both target and non-target actions. If a pair of features with indices i, j passes this filter, then its perturbation direction is determined by $d_{perturbation}$. To find an adversarial example, our method iteratively finds a pair of input features that has the highest saliency score and following:
\begin{equation}
\begin{split}
& \boldsymbol{X}_{adv, 0}  = \boldsymbol{X}\\
& \boldsymbol{X}_{adv, t_{iter}}[i] = \boldsymbol{X}_{adv, t_{iter}-1}[i] + d_{perturbation} \times \theta \\
& \boldsymbol{X}_{adv, t_{iter}}[j] = \boldsymbol{X}_{adv, t_{iter}-1}[j] + d_{perturbation} \times \theta
\end{split}
\end{equation}
where the $\theta$ is the step size of the perturbation, and $t_{iter}$ is the number of iterations. For each time step in a c-MARL game, this method is invoked iteratively until a successful adversarial example is found or the $t_{iter}$ reaches max iteration.

The method finds a saliency score for each pair of feature gradients and creates a bigger saliency map. By considering two feature gradients at the same time, we obtain more accurate estimations for the output Q-value contour of the model, which further helps us find better options to explore the output space. This method differs from the original JSMA implementation in that it considers both the positive target action Q-value gradient, as well as the negative ones (in which case the $d_{perturbation}$ is negative). We refer to this method as JSMA, using the notation d-JSMA to denote dynamic $\theta$, and JSMA for a fixed $\theta$. To further improve the efficiency, more than two feature points can be considered at the same time, however, the calculation complexity is $O(n^d)$ ($d$ is the number of features). 

2. \textbf{Dynamic $\boldsymbol{\theta}$}: JSMA has a hyperparameter $\theta$ which controls the perturbation size for each attack iteration. In our problem setup, we found that a fixed $\theta$ does not perform well, and a large $\theta$ does not always outperform a small $\theta$. One hypothesis for this behavior is due to the fast-changing gradient in the output contour surface. To achieve better attack performance, we propose using dynamic $\theta$ in the d-JSMA method. For each time step in the c-MARL game, we start with a low $\theta$ value for our attack, and if the attack is not successful, we increase the size of $\theta$ and retry the attack until we reach the maximum $\theta$ value.

\section{Experiments}\label{ex}

\subsection{Experimental Setup:}\label{exp}

We evaluate our attacks on the StarCraft Multi-Agent Challenge (SMAC) \cite{samvelyan2019starcraft}. StarCraft II is a real-time strategy game used widely as a benchmark for RL research. We used the "2s3z" SMAC map (2 "Stalkers", 3 "Zealots" per team) for our scenarios. 
Our ally team is a c-MARL system whose goal is to defeat the opponents.  
For our attack, one of the ally Stalkers is the victim, and the rest of the allies use their original policy, trained in a centralized manner to achieve high team win rate. More details on the features and actions can be found in \cite{samvelyan2019starcraft}.

\subsection{Results and Discussion:}

\subsubsection{Learning adversarial policy}

To learn an adversarial policy for the victim agent, we applied the methods from~\ref{sec:direct_control}. We trained an adversarial policy for the victim to select sub-optimal actions that minimize the total reward. To evaluate the performance of the policy, we directly control the action of the victim based on the output of our adversarial policy. We ran each attack method for 500 games and presented the results 
in table~\ref{tab:1}. From the results, OW and OWR have the highest negative impact on team reward and win rate, with 100\% loss rate, making these two methods the most efficient. 

\begin{table}[ht]
    \centering
    \caption{Learning adversarial policies to select the victim's target action.}
    \scalebox{0.97}{
    \begin{tabular}[t]{lcc}
        \toprule
        &Average Reward & Team Win Rate\\
        \midrule
        Baseline   & 20.00 & 99.80\%\\
        Random &   11.10  & 3.40\%\\
        QMIX Worst    & 10.58 & 1.41\%\\
        Local Worst & 10.52 & 1.00\%\\
        \textbf{Optimized Worst (OW)}&   9.39  & \textbf{0\%}\\
        \textbf{Optimized Worst with Reg (OWR)} & \textbf{9.35} & \textbf{0\%} \\ 
        \bottomrule
    \end{tabular}}
    \label{tab:1}
\end{table}

\subsubsection{d-JSMA methods}
We evaluate d-JSMA with fixed attack step size $\theta$ (0.1 - 0.9), and a dynamic one. We run 100 games for each configuration and report their average results. 
For the target action, we combine d-JSMA with LW, OW and OWR.  

\begin{figure}
    \centering
    \begin{subfigure}{0.5\textwidth} 
        \centering
        \includegraphics[width=1\linewidth]{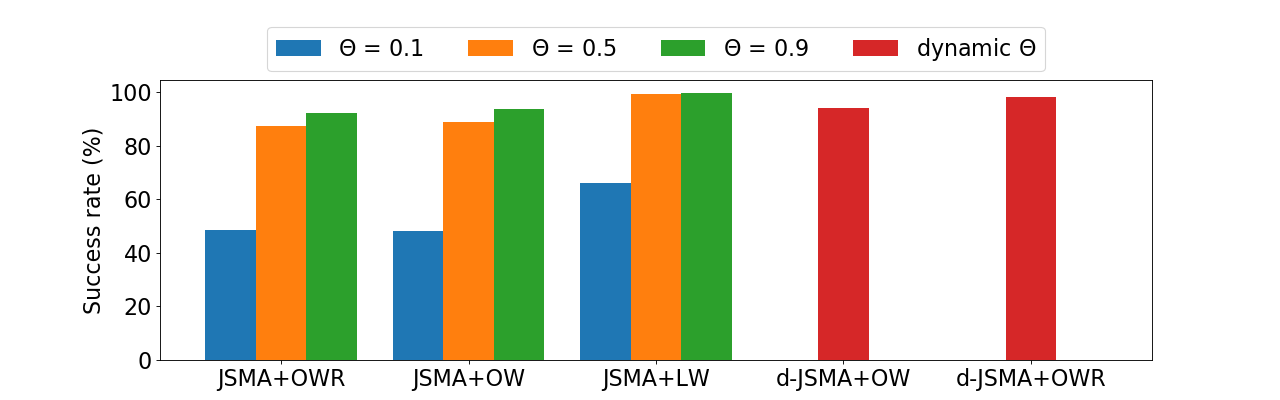}
        \caption{JSMA-based Methods Target Action Success Rate}
    \end{subfigure}
    \begin{subfigure}{0.5\textwidth}
        \centering
        \includegraphics[width=\linewidth]{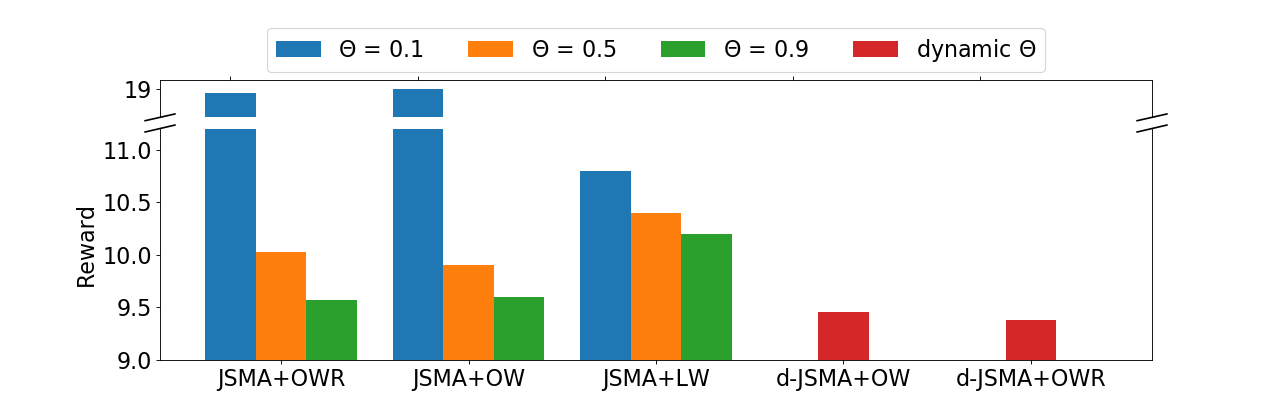}
        \caption{JSMA-based Methods Total Reward}
    \end{subfigure}

    \caption{JSMA-based Methods Evaluation under Different $\theta$}
    \label{fig:jsma_eval}
\end{figure}

\begin{table}[ht]
\centering
\caption{Team Win Rate of JSMA-based Methods.}
\begin{tabular}[t]{c|c|ccc}
\toprule
&$\theta$ & LW & OW & OWR  \\
\midrule
 & 0.1   & 2.0\% & 83.0\% & 81.0\%\\
JSMA & 0.5 &   0.0\%  & 1.0\% & 1.0\%\\
 & 0.9 &   0.0\% & 0.0\% & 0.0\%\\
\midrule
d-JSMA & dynamic & 0.00\% & 0.00\% & 0.00\% \\
\bottomrule
\end{tabular}
\label{table:jsma_winrate}
\end{table}%

Figure \ref{fig:jsma_eval} shows the total reward, and target action success rate for d-JSMA and JSMA with different $\theta$ values. Table \ref{table:jsma_winrate} shows the team win rate for both methods. The results show that as we increase the input perturbation attack budget, the total team reward goes down as expected. At a low budget, JSMA+LW has relatively better attack performance. If a higher budget is available, then d-JSMA+OWR is able to achieve the lowest reward. Moreover, d-JSMA+OWR achieves better performance than JSMA+LW, even if the target action success rate is slightly lower, hence showing that the underlying OWR action selection method is more efficient. In this experiment, all the features are assumed to be continuous. Considering some of the features in the SMAC benchmark are discrete in nature, we have also conducted experiments by only perturbing the continuous features, and we did not notice a major performance difference.

\subsubsection{Overall performance \& budget comparison of all methods}
We evaluate the overall performance of the two-step attack methods that combine the learning adversarial policy methods and the adversarial example generation methods. Table \ref{table:budget} shows the best attack performance for each method and the total team reward reduction achieved. Based on the results, d-JSMA+OWR achieves the best attack performance. It achieves 10.62 reward reduction, which is close to the one achieved by directly controlling the victim with OWR (10.65). Moreover, d-JSMA+OWR outperforms d-JSMA+OW, so we conclude that adding regularization to OW as described in~\ref{sec:atatcks}, does provide a benefit when combining the two attack steps. 

\begin{figure}[t]
    \begin{subfigure}[b]{0.24\textwidth}
        \includegraphics[width=\linewidth]{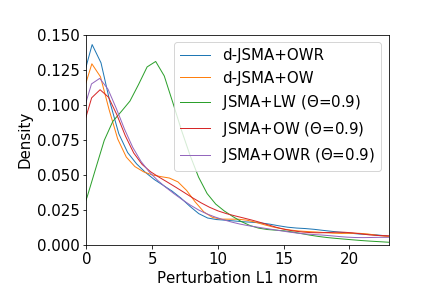}
        \caption{Perturbation distribution from 0 to 20}
    \end{subfigure}
    \begin{subfigure}[b]{0.24\textwidth}
        \includegraphics[width=\linewidth]{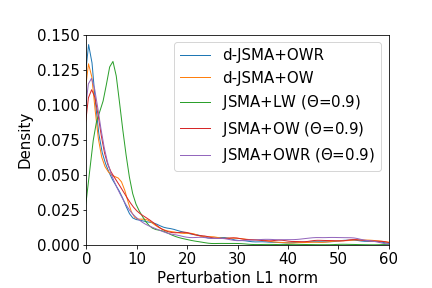}
        \caption{Perturbation distribution from 0 to 70}
    \end{subfigure}
    \caption{Distribution of required perturbations for different attack methods.}
    \label{fig:budget_dist}
\end{figure}

In table \ref{table:budget}, we also show the evaluation for the perturbation budget. d-JSMA+OWR requires a lower budget than most other JSMA methods, while achieving the best total reward reduction. Figure \ref{fig:budget_dist} shows the perturbation distribution, where the d-JSMA+OWR method has the highest peak densities close to 0, showing that our regularization method effectively reduces the required perturbation. d-JSMA+OWR also has a lower density than d-JSMA+OW after 5 L1 norm, confirming regularization's effectiveness. Finally, d-JSMA-based methods generally have higher density in the lower perturbation range compared to JSMA methods, thus can achieve good performance with a relatively low budget.

\begin{table}[t]
\centering
\caption{Average perturbation required (in terms of L1 norm) v.s. total reward reduction for JSMA and d-JSMA methods.}
\begin{tabular}[t]{c|c|ccc}
\toprule
 &Total Reward & Average L1 Norm\\ & Reduction & Perturbation\\
\midrule

d-JSMA+OWR & \textbf{10.62} &  8.33  \\
d-JSMA+OW & 10.54 & 9.01  \\
JSMA+LW ($\theta=0.9$) & 9.78 & \textbf{6.34}  \\
JSMA+OW ($\theta=0.9$) & 10.42 & 9.62  \\
JSMA+OWR ($\theta=0.9$) & 10.43 & 9.80  \\
it-FGSM+LW & 8.36 & 20.37\\

\bottomrule
\end{tabular}
\label{table:budget}
\end{table}%

\subsection{Attack Strategy in StarCraft benchmark environment}
In this section, we examine the learned attack strategy and behaviors. Using LW or QMIX worst action selection policy (Figure~\ref{subfig:QMIX}) results 
in the victim walking away from the opponent's range of sight---staying hidden, even after its team is defeated. The episode ends with the victim alive. 

\subsubsection{Agent behaviour under OW and OWR adversarial policies}

\begin{figure}[t]
    \begin{subfigure}[t]{0.24\textwidth}
        \includegraphics[width=\linewidth]{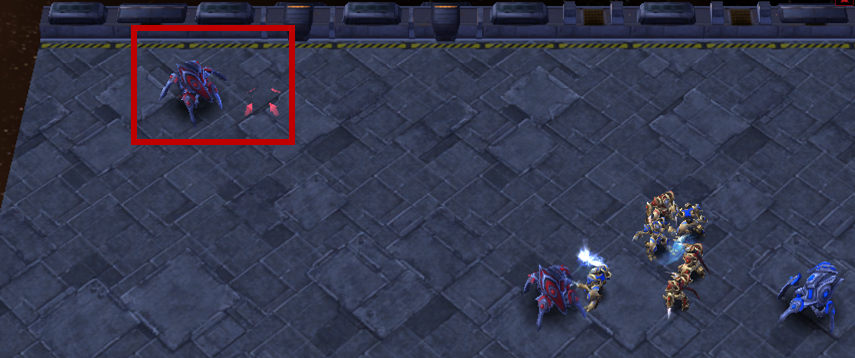}
        \caption{QMIX worst, local worst adversarial policy.}
        \label{subfig:QMIX}
    \end{subfigure}
    \begin{subfigure}[t]{0.24\textwidth}
        \includegraphics[width=\linewidth]{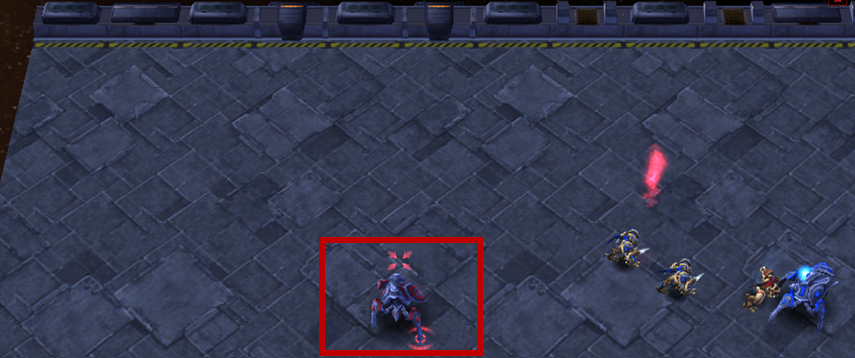}
        \caption{OW, OWR adversarial policy and combined two-step attack (d-JSMA + OWR)}
        \label{subfig:OW}
    \end{subfigure}
    \caption{Screenshots from replayed episodes. Victim agent shown in rectangle.}
    \label{fig:screen}
\end{figure}

At the beginning of an episode, the victim moves away from its team immediately, hiding until its team is almost defeated (as in~\ref{subfig:QMIX}). Then, it moves towards the opponents to be defeated, as in~\ref{subfig:OW}. Once within their sight range, most often it backs away, resulting in it being defeated. We have even observed the victim luring some of its team at times. 

\subsubsection{Two-step attack - d-JSMA+OWR}

By only perturbing the victim's observation, our two-step attack can lure the victim to behave similarly as being directly controlled by the OWR policy. We can see the victim occasionally select sub-optimal actions, but the attack method is able to quickly recover.

\section{Discussion on Defense}
\label{sec:defenses}
To improve robustness of c-MARL systems, adversarial manipulations of both team agents and the environment need to be taken into account during training. Being able to isolate the impact of malicious agents is crucial in achieving robustness. One possible defense mechanism is to have each agent estimate action values or reward function for other agents, and use the estimate to potentially identify malicious agents. Techniques from Inverse Reinforcement Learning and model-based RL can be applied here. Another possible defense method involves formulating all agents as potential adversaries during the MARL centralized training, so that an agent can react better to adversarial actions during execution.

\section{Related Work}\label{related}
Prior work has demonstrated that RL is vulnerable to input perturbation attacks, where the attacker modifies the agent’s observation with the goal of degrading its performance; for example, using the FGSM~\cite{goodfellow2014explaining} to attack three RL networks~\cite{huang2017adversarial}. Other attacks reduce 
 the number of adversarial examples needed to decrease the agent's reward~\cite{lin2017tactics} or trigger misbehavior of the agent after a delay~\cite{zhao2019blackbox}. 
The learning process of DQN itself can be attacked~\cite{behzadan2017vulnerability}, by constructing a replica model of the victim and transferring adversarial examples from the FGSM and JSMA~\cite{papernot2016limitations} techniques to craft adversarial examples. 
None of this work studies c-MARL, and the effects of cooperation. 

The closest work to ours is perhaps Gleave et al.~\cite{gleave2019adversarial}. They focus on attacks in the competitive multi-agent setting whereas we instead consider cooperative teams of agents. Furthermore, they assume full control over one of the agents (i.e., they were able to directly control the agent's actions), whereas we only perturb the agent's environment. 

\section{Conclusions}\label{conc}
We presented the first robustness analysis of cooperative teams of RL agents. Using the StarCraft II multi-agent environment as a benchmark, we proposed a two-step attack that combines 1) learning an adversarial policy and 2) gradient-based adversarial example generation. We demonstrated the possibility of controlling the total team reward by attacking a single agent. We hope our attack method can serve as a tool to assist evaluating and validating the robustness of c-MARL.

\bibliographystyle{IEEEtran}
\bibliography{IEEEabrv,ref}

\section*{Acknowledgments}

\noindent This research was supported in part by CIFAR.

\end{document}